\documentclass[10pt,twocolumn,letterpaper]{article}
\pdfoutput=1

\usepackage[pagenumbers]{cvpr} 

\usepackage{graphicx}
\usepackage{amsmath}
\usepackage{amssymb}
\usepackage{booktabs}
\usepackage{multirow}
\usepackage{booktabs}

\usepackage{algorithm}
\usepackage{algorithmicx}
\usepackage{mathtools}
\usepackage{adjustbox}
\usepackage{import}
\usepackage[noend]{algpseudocode}
\usepackage{caption}
\usepackage{subcaption}
\usepackage[dvipsnames]{xcolor}
\usepackage{balance}
\makeatletter
\@namedef{ver@everyshi.sty}{}
\makeatother
\usepackage{tikz}
\usetikzlibrary{tikzmark}
\usepackage{balance}

\newlength\savewidth

\renewcommand{\paragraph}[1]{\vspace{1.25mm}\noindent\textbf{#1}}

\newcommand{\app}{\raise.17ex\hbox{$\scriptstyle\sim$}}
\usepackage[pagebackref=false, breaklinks=true, letterpaper=true, colorlinks, urlcolor=urlcolor,
citecolor=citecolor, linkcolor=linkcolor, bookmarks=false]{hyperref}
\definecolor{citecolor}{HTML}{0071BC}
\definecolor{linkcolor}{HTML}{ED1C24}
\definecolor{urlcolor}{HTML}{0e84e6}

\usepackage[capitalize]{cleveref}
\crefname{section}{Sec.}{Secs.}
\Crefname{section}{Section}{Sections}
\Crefname{table}{Table}{Tables}
\crefname{table}{Tab.}{Tabs.}

\def\cvprPaperID{7069} 
\def\confName{CVPR}
\def\confYear{2023}

\newcommand{\methodnameA}{\mbox{LART-pose}}
\newcommand{\methodnameB}{\mbox{LART}}

\begin{document}

\title{On the Benefits of 3D Pose and Tracking for Human Action Recognition}

\author{
\kern-7mm Jathushan Rajasegaran$^{1,2}$, Georgios Pavlakos$^{1}$, Angjoo Kanazawa$^{1}$, Christoph Feichtenhofer$^{2}$, Jitendra Malik$^{1,2}$\\
\kern-7mm $^{1}$UC Berkeley, $^{2}$Meta AI, FAIR
}


\maketitle




 \pdfoutput=1
\begin{abstract}
In this work we study the benefits of using tracking and 3D poses for action recognition. To achieve this, we take the Lagrangian view on analysing actions over a trajectory of human motion rather than at a fixed point in space. Taking this stand allows us to use the tracklets of people to predict their actions. In this spirit, first we show the benefits of using 3D pose to infer actions, and study person-person interactions. Subsequently, we propose a Lagrangian Action Recognition model by fusing 3D pose and contextualized appearance over tracklets. To this end, our method achieves state-of-the-art performance on the AVA v2.2 dataset on both pose only settings and on standard benchmark settings. When reasoning about the action using only pose cues, our pose model achieves \textbf{+10.0 mAP} gain over the corresponding state-of-the-art while our fused model has a gain of \textbf{+2.8 mAP} over the best state-of-the-art model. Our best model achieves 45.1 mAP on AVA 2.2 dataset. Code and results are available at: \href{https://brjathu.github.io/LART}{https://brjathu.github.io/LART}
\end{abstract}
 \pdfoutput=1
\section{Introduction}
\label{sec:introduction}

In fluid mechanics, it is traditional to distinguish between the Lagrangian and Eulerian specifications of the flow field. Quoting the~\href{https://en.wikipedia.org/wiki/Lagrangian_and_Eulerian_specification_of_the_flow_field}{Wikipedia entry},
\emph{``Lagrangian specification of the flow field is a way of looking at fluid motion where the observer follows an individual fluid parcel as it moves through space and time. Plotting the position of an individual parcel through time gives the pathline of the parcel. This can be visualized as sitting in a boat and drifting down a river. The Eulerian specification of the flow field is a way of looking at fluid motion that focuses on specific locations in the space through which the fluid flows as time passes. This can be visualized by sitting on the bank of a river and watching the water pass the fixed location.''} 

These concepts are very relevant to how we analyze videos of human activity. In the Eulerian viewpoint, we would focus on feature vectors at particular locations, either $(x,y)$ or $(x,y,z)$, and consider evolution over time while staying fixed in space at the location. In the Lagrangian viewpoint, we would track, say a person over space-time and track the associated feature vector across space-time.

While the older literature for activity recognition \eg, \cite{efros2003recognizing,Wang2011action,gkioxari2015finding} typically adopted the Lagrangian viewpoint, ever since the advent of neural networks based on 3D space-time convolution, \eg,~\cite{tran2015learning}, the Eulerian viewpoint became standard in state-of-the-art approaches such as  SlowFast Networks~\cite{feichtenhofer2019slowfast}. Even after the switch to transformer architectures~\cite{vaswani2017attention, fan2021multiscale} the Eulerian viewpoint has persisted. This is noteworthy because the tokenization step for transformers gives us an opportunity to freshly examine the question, \emph{``What should be the counterparts of words in video analysis?''}. Dosovitskiy~\etal~\cite{dosovitskiy2020image} suggested that image patches were a good choice, and the continuation of that idea to video suggests that spatiotemporal cuboids would work for video as well.

\begin{figure*}[!ht]
    \centering
    \includegraphics[width=0.86\textwidth,trim={0 0 0 0},clip]{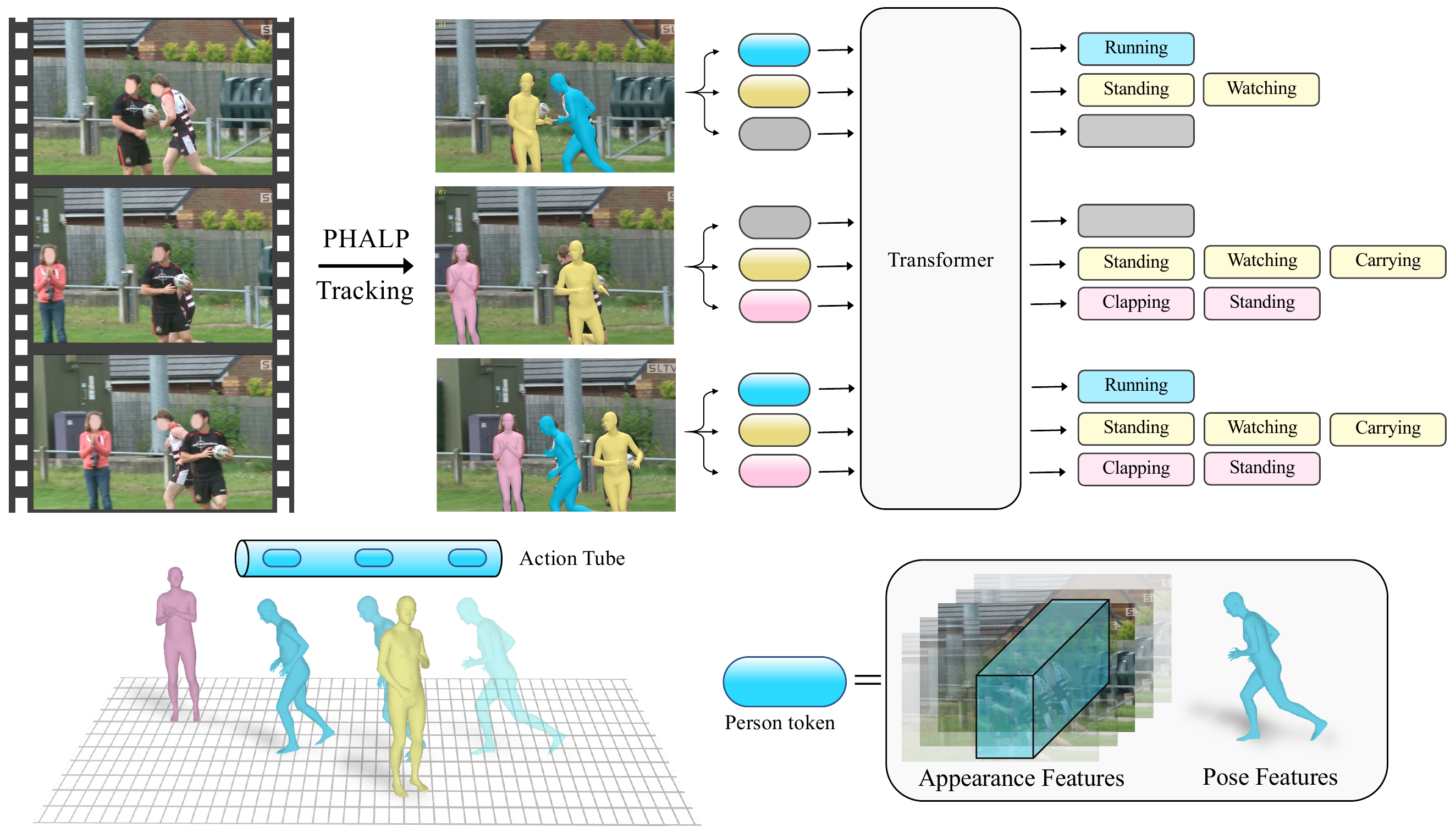}
    \caption{\textbf{Overview of our method:} Given a video, first, we track every person using a tracking algorithm (\emph{e.g.} PHALP~\cite{rajasegaran2022tracking}). Then every detection in the track is tokenized to represent a human-centric vector (\emph{e.g. pose, appearance}). To represent 3D pose we use SMPL~\cite{loper2015smpl} parameters and estimated 3D location of the person, for contextualized appearance we use MViT~\cite{fan2021multiscale} (pre-trained on MaskFeat~\cite{wei2022masked}) features. Then we train a transformer network to predict actions using the tracks. Note that, at the second frame we do not have detection for the \tikzmarknode[draw,inner sep=2pt,rounded corners,fill=Cerulean!30]{A}{blue person}, at these places we pass a \tikzmarknode[draw,inner sep=2pt,rounded corners,fill=gray!30]{A}{mask token} to in-fill the missing detections. }
    \label{fig:model}
\end{figure*}

On the contrary, in this work we take the Lagrangian viewpoint for analysing human actions. This specifies that we reason about the \emph{trajectory of an entity} over time. Here, the entity can be low-level, \eg, a pixel or a patch, or high-level, \eg, a person. Since, we are interested in understanding human actions, we choose to operate on the level of \emph{``humans-as-entities''}. To this end, we develop a method that processes trajectories of people in video and uses them to recognize their action. We recover these trajectories by capitalizing on a recently introduced 3D tracking method PHALP~\cite{rajasegaran2022tracking} and HMR 2.0~\cite{4DHUMANS}. As shown in Figure~\ref{fig:model} PHALP recovers person tracklets from video by lifting people to 3D, which means that we can both link people over a series of frames and get access to their 3D representation. Given these 3D representations of people (\ie, 3D pose and 3D location), we use them as the basic content of each token. This allows us to build a flexible system where the model, here a transformer, takes as input tokens corresponding to the different people with access to their identity, 3D pose and 3D location. Having 3D location of the people in the scene allow us to learn interaction among people. Our model relying on this tokenization can benefit from 3D tracking and pose, and outperforms previous baseline that only have access to pose information~\cite{choutas2018potion,shah2022pose}.

While the change in human pose over time is a strong signal, some actions require more contextual information about the appearance and the scene. Therefore, it is important to also fuse pose with appearance information from humans and the scene, coming directly from pixels. To achieve this, we also use the state-of-the-art models for action recognition~\cite{fan2021multiscale,li2021improved} to provide complementary information from the contextualized appearance of the humans and the scene in a Lagrangian framework. Specifically, we densely run such models over the trajectory of each tracklet and record the contextualized appearance features localized around the tracklet. As a result, our tokens include explicit information about the 3D pose of the people and densely sampled appearance information from the pixels, processed by action recognition backbones~\cite{fan2021multiscale}. Our complete system outperforms the previous state of the art by a large margin of \textbf{2.8 mAP}, on the challenging AVA v2.2 dataset. 

Overall, our main contribution is introducing an approach that highlights the effects of tracking and 3D poses for human action understanding. To this end, in this work, we propose a \textbf{L}agrangian \textbf{A}ction \textbf{R}ecognition with \textbf{T}racking (\textbf{\methodnameB}) approach, which utilizes the tracklets of people to predict their action. Our baseline version leverages tracklet trajectories and 3D pose representations of the people in the video to outperform previous baselines utilizing pose information. Moreover, we demonstrate that the proposed Lagrangian viewpoint of action recognition can be easily combined with traditional baselines that rely only on appearance and context from the video, achieving significant gains compared to the dominant paradigm.

 \pdfoutput=1
\section{Related Work}
\label{sec:related_work}

\paragraph{Recovering humans in 3D:} A lot of the related work has been using the SMPL human body model~\cite{loper2015smpl} for recovering 3D humans from images. Initially, the related methods were relying on optimization-based approaches, like SMPLify~\cite{bogo2016keep}, but since the introduction of the HMR~\cite{kanazawa2018end}, there has been a lot of interest in approaches that can directly regress SMPL parameters~\cite{loper2015smpl} given the corresponding image of the person as input. Many follow-up works have improved upon the original model, estimating more accurate pose~\cite{kolotouros2019learning} or shape~\cite{choutas2022accurate}, increasing the robustness of the model~\cite{pavlakos2022human}, incorporating side information~\cite{kocabas2021spec,kolotouros2021probabilistic}, investigating different architecture choices~\cite{kocabas2021pare,zhang2021pymaf}, etc.

While these works have been improving the basic single-frame reconstruction performance, there have been parallel efforts toward the temporal reconstruction of humans from video input. The HMMR model~\cite{kanazawa2019learning} uses a convolutional temporal encoder on HMR image features~\cite{kanazawa2018end} to reconstruct humans over time. Other approaches have investigated recurrent~\cite{kocabas2020vibe} or transformer~\cite{pavlakos2022human} encoders. Instead of performing the temporal pooling on image features, recent work has been using the SMPL parameters directly for the temporal encoding~\cite{baradel2021leveraging,rempe2021humor}.

One assumption of the temporal methods in the above category is that they have access to tracklets of people in the video. This means that they rely on tracking methods, most of which operate on the 2D domain~\cite{bergmann2019tracking,fang2017rmpe,meinhardt2022trackformer, xiu2018pose} and are responsible for introducing many errors. To overcome this limitation, recent work~\cite{rajasegaran2021tracking,rajasegaran2022tracking} has capitalized on the advances of 3D human recovery to perform more robust identity tracking from video. More specifically, the PHALP method of Rajasegaran~\etal~\cite{rajasegaran2022tracking} allows for robust tracking in a variety of settings, including in the wild videos and movies. Here, we make use of the PHALP system to discover long tracklets from large-scale video datasets. This allows us to train our method for recognizing actions from 3D pose input.

\paragraph{Action Recognition:} Earlier works on action recognition relied on hand-crafted features such as HOG3D~\cite{klaser2008spatio}, Cuboids~\cite{dollar2005behavior} and Dense Trajectories~\cite{Wang2011action,wang2013action}. After the introduction of deep learning, 3D convolutional networks became the main backbone for action recognition~\cite{taylor2010convolutional, tran2015learning, carreira2017quo}. However, the 3D convolutional models treat both space and time in a similar fashion, so to overcome this issue, two-stream architectures were proposed~\cite{simonyan2014two}. In two-steam networks, one pathway is dedicated to motion features, usually taking optical flow as input. This requirement of computing optical flow makes it hard to learn these models in an end-to-end manner. On the other hand, SlowFast networks~\cite{feichtenhofer2019slowfast} only use video streams but at different frame rates, allowing it to learn motion features from the fast pathway and lateral connections to fuse spatial and temporal information. Recently, with the advancements in transformer architectures, there has been a lot of work on action recognition using transformer backbones~\cite{neimark2021video, bertasius2021space, arnab2021vivit, fan2021multiscale}. 

While the above-mentioned works mainly focus on the model architectures for action recognition, another line of work investigates more fine-grained relationships between actors and objects~\cite{wang2018non, wang2018videos, sun2019relational, zhang2019structured}. Non-local networks~\cite{wang2018non} use self-attention to reason about entities in the video and learn long-range relationships. ACAR~\cite{pan2021actor} models actor-context-actor relationships by first extracting actor-context features through pooling in bounding box region and then learning higher-level relationships between actors. Compared to ACAR, our method does not explicitly design any priors about actor relationships, except their track identity.

Along these lines, some works use the human pose to understand the action~\cite{choutas2018potion, yan2019pa3d, weinzaepfel2021mimetics, shah2022pose, varol2021synthetic}. PoTion~\cite{choutas2018potion} uses a keypoint-based pose representation by colorizing the temporal dependencies. Recently, JMRN~\cite{shah2022pose} proposed a joint-motion re-weighting network to learn joint trajectories separately and then fuse this information to reason about inter-joint motion. While these works rely on 2D key points and design-specific architectures to encode the representation, we use more explicit 3D SMPL parameters.
 \pdfoutput=1
\section{Method}
\label{sec:method}

Understanding human action requires interpreting multiple sources of information~\cite{keestra2015understanding}. These include head and gaze direction, human body pose and dynamics, interactions with objects or other humans or animals, the scene as a whole, the activity context (e.g. immediately preceding actions by self or others), and more. Some actions can be recognized by pose and pose dynamics alone, as demonstrated by Johansson \emph{et al}~\cite{johansson1973visual} who showed that people are remarkable at recognizing \emph{walking}, \emph{running}, \emph{crawling} just by looking at moving point-lights. However, interpreting complex actions requires reasoning with multiple sources of information e.g. to recognize that someone is slicing a tomato with a knife, it helps to see the knife and the tomato.

There are many design choices that can be made here. Should one use ``disentangled" representations, with elements such as pose, interacted objects, etc, represented explicitly in a modular way? Or should one just input video pixels into a large capacity neural network model and rely on it to figure out what is discriminatively useful? In this paper, we study two options: a) human pose reconstructed from an HMR model~\cite{kanazawa2018end, 4DHUMANS} and b) human pose with contextual appearance as computed by an MViT model~\cite{fan2021multiscale}.

Given a video with number of frames $T$, we first track every person using PHALP~\cite{rajasegaran2022tracking}, which gives us a unique identity for each person over time. Let a person $i \in [1,2,3,...n]$ at time $t \in [1,2,3, ... T]$ be represented by a person-vector $\mathcal{\textbf{H}}_t^i$. Here $n$ is the number of people in a frame. This person-vector is constructed such that, it contains human-centric representation $\mathcal{\mathbf{P}}_t^i$ and some contextualized appearance information $\mathcal{\mathbf{Q}}_t$.
\begin{equation}
    \mathcal{\mathbf{H}}_t^i = \{\mathcal{\mathbf{P}}_t^i, \mathcal{\mathbf{Q}}_t^i\}.
    \vspace{-4pt}
\end{equation}

Since we know the identity of each person from the tracking, we can create an action-tube~\cite{gkioxari2015finding} representation for each person. Let $\mathcal{\mathbf{\Phi}}_i$ be the action-tube of person $i$, then this action-tube contains all the person-vectors over time.
\begin{equation}
    \mathcal{\mathbf{\Phi}}_i = \{\mathcal{\mathbf{H}}_1^i, \mathcal{\mathbf{H}}_2^i, \mathcal{\mathbf{H}}_3^i, ..., \mathcal{\mathbf{H}}_T^i\}.
    \vspace{-4pt}
\end{equation}
Given this representation, we train our model \textbf{\methodnameB} to predict actions from action-tubes (tracks). In this work we use a vanilla transformer~\cite{vaswani2017attention} to model the network $\mathcal{F}$, and this allow us to mask attention, if the track is not continuous due to occlusions and failed detections etc. Please see the Appendix for more details on network architecture.
\begin{equation}
    \mathcal{F}\big( \mathbf{\Phi}_1, \mathbf{\Phi}_2, ...,  \mathbf{\Phi}_i, ..., \mathbf{\Phi}_n ; \Theta \big) = \widehat{Y_i}.
\end{equation}
Here, $\Theta$ is the model parameters, $\widehat{Y_i} = \{y_1^i, y_2^i, y_3^i, ... , y_T^i\}$ is the predictions for a track, and $y_t^i$ is the predicted action of the track $i$ at time $t$. The model can use the actions of others for reasoning when predicting the action for the person-of-interest $i$. Finally, we use binary cross-entropy loss to train our model and measure mean Average Precision (mAP) for evaluation.

\subsection{Action Recognition with 3D Pose}
\label{sec:method_pose}

In this section, we study the effect of human-centric pose representation on action recognition. To do that, we consider a person-vector that only contains the pose representation, $\mathcal{\mathbf{H}}_t^i = \{\mathcal{\mathbf{P}}_t^i\}$. While, $\mathcal{\mathbf{P}}_t^i$ can in general contain any information about the person, in this work train a pose only model \textbf{\methodnameA} which uses 3D body pose of the person based on the SMPL~\cite{loper2015smpl} model. This includes the joint angles of the different body parts, $\mathcal{\theta}_t^i \in \mathcal{R}^{23\times3\times3}$ and is considered as an amodal representation, which means we make a prediction about all body parts, even those that are potentially occluded/truncated in the image. Since the global body orientation $\psi_{t}^i \in \mathcal{R}^{3\times3}$ is represented separately from the body pose, our body representation is invariant to the specific viewpoint of the video. In addition to the 3D pose, we also use the 3D location $L_t^i$ of the person in the camera view (which is also predicted by the PHALP model~\cite{rajasegaran2022tracking}). This makes it possible to consider the relative location of the different people in 3D. More specifically, each person is represented as,
\begin{equation}
    \mathcal{\mathbf{H}}_t^i = \mathcal{\mathbf{P}}_t^i = \{\theta_t^i, \psi_t^i, L_t^i \}. 
\end{equation}

Let us assume that there are $n$ tracklets $\{ \mathbf{\Phi}_1, \mathbf{\Phi}_2, \mathbf{\Phi}_3, ..., \mathbf{\Phi}_n \}$ in a given video. To study the action of the tracklet $i$, we consider that person $i$ as the person-of-interest and having access to other tracklets can be helpful to interpret the person-person interactions for person $i$. Therefore, to predict the action for all $n$ tracklets we need to make $n$ number of forward passes. If person $i$ is the person-of-interest, then we randomly sample $N-1$ number of other tracklets and pass it to the model $\mathcal{F}(;\Theta)$ along with the $\mathbf{\Phi}_i$.
\begin{equation}
    \mathcal{F}(\mathcal{\mathbf{\Phi}}_i, \{\mathcal{\mathbf{\Phi}}_j | j \in [N]\}; \Theta) = \widehat{Y_i}
\end{equation}

Therefore, the model sees $N$ number of tracklets and predicts the action for the main (person-of-interest) track. To do this, we first tokenize all the person-vectors, by passing them through a linear layer and project it in $f_{proj}(\mathbf{\mathcal{H}}_t^i) \in \mathcal{R}^d$ a $d$ dimensional space. Afterward, we add positional embeddings for a) time, b) tracklet-id. For time and tracklet-id we use 2D sine and cosine functions as positional encoding~\cite{wang2021translating}, by assigning person $i$ as the zero$^\text{th}$ track, and the rest of the tracklets use tracklet-ids $\{1,2,3,...,N-1\}$. 
\begin{align*}
    \textit{PE}(t, i, 2r) = \sin(t/10000^{4r/d}) \\
    \textit{PE}(t, i, 2r+1) = \cos(t/10000^{4r/d}) \\
    \textit{PE}(t, i, 2s+D/2) = \sin(i/10000^{4s/d}) \\
    \textit{PE}(t, i, 2s+D/2+1) = \cos(i/10000^{4s/d})
\end{align*}
Here, $t$ is the time index, $i$ is the track-id, $r,s \in [0, d/2)$ specifies the dimensions and $D$ is the dimensions of the token. 

After adding the position encodings for time and identity, each person token is passed to the transformer network. The $(t+i\times N)^{th}$ token is given by,
\begin{equation}
    token_{(t+i\times N)} = f_{proj}(\mathbf{\mathcal{H}}_t^i) + \textit{PE}(t, i, :) 
\end{equation}

Our person of interest formulation would allow us to use other actors in the scene to make better predictions for the main actor. When there are multiple actors involved in the scene,  knowing one person's action could help in  predicting another's action. Some actions are correlated among the actors in a scene (e.g. \textit{dancing}, \textit{fighting}), while in some cases, people will be performing reciprocal actions (e.g. \textit{speaking} and \textit{listening}). In these cases knowing one person's action would help in predicting the other person's action with more confidence.

\subsection{Actions from Appearance and 3D Pose}
\label{sec:method_pose_apperance}

While human pose plays a key role in understanding actions, more complex actions require reasoning about the scene and context. Therefore, in this section, we investigate the benefits of combining pose and contextual appearance features for action recognition and train model \textbf{\methodnameB} to benefit from 3D poses and appearance over a trajectory. For every track, we run a 2D action recognition model (\ie MaskFeat~\cite{wei2022masked} pretrained MViT~\cite{fan2021multiscale}) at a frequency $f_s$ and store the feature vectors before the classification layer. For example, consider a track $\mathbf{\Phi}_i$, which has detections $\{D_1^i, D_2^i, D_3^i, ..., D_T^i\}$. We get the predictions form the 2D action recognition models, for the detections at $\{t, t+f_{FPS}/f_{s}, t+ 2 f_{FPS}/f_{s}, ...\}$. Here, $f_{FPS}$ is the rate at which frames appear on the screen. Since these action recognition models capture temporal information to some extent, $\mathbf{Q}_{t-f_{FPS}/2f_s}^i$ to $\mathbf{Q}_{t+f_{FPS}/2f_s}^i$ share the same appearance features. Let's assume we have a pre-trained action recognition model $\mathcal{A}$, and it takes a sequence of frames and a detection bounding box at mid-frame, then the feature vectors for $\mathbf{Q}_{t}^i$ is given by:
\begin{equation*}
    \mathcal{A}\big(D_t^i, \{I\}_{t-M}^{t+M}\big) = \mathbf{U}_t^i
\end{equation*}
Here, $\{I\}_{t-M}^{t+M}$ is the sequence of image frames, $2M$ is the number of frames seen by the action recognition model, and $\mathbf{U}_t^i$ is the contextual appearance vector. Note that, since the action recognition models look at the whole image frame, this representation implicitly contains information about the scene and objects and movements. However, we argue that human-centric pose representation has orthogonal information compared to feature vectors taken from convolutional or transformer networks. For example, the 3D pose is a geometric representation while $\mathbf{U}_t^i$ is more photometric, the SMPL parameters have more priors about human actions/pose and it is amodal while the appearance representation is learned from raw pixels. Now that we have both pose-centric representation and appearance-centric representation in the person vector $\mathcal{\mathbf{H}}_t^i$:
\begin{align}
    \mathcal{\mathbf{H}}_t^i  &= \{\underbrace{\theta_t^i, \psi_t^i, L_t^i}_{\mathbf{P}_t^i}, \underbrace{\mathbf{U}_t^i}_{\mathbf{Q}_t^i}\}
\end{align}
So, each human is represented by their 3D pose, 3D location, and with their appearance and scene content. We follow the same procedure as discussed in the previous section to add positional encoding and train a transformer network $\mathcal{F}(\Theta)$ with pose+appearance tokens. 
\section{Experiments}
\label{sec:experiments}

We evaluate our method on AVA~\cite{gu2018ava} in various settings. AVA~\cite{gu2018ava} poses an action detection problem, where people are localized in a spatio-temporal volume with action labels. It provides annotations at 1Hz, and each actor will have 1 pose action, up to 3 person-object interactions (optional), and up to 3 person-person interaction (optional) labels. For the evaluations, we use AVA v2.2 annotations and follow the standard protocol as in ~\cite{gu2018ava}. We measure mean average precision (mAP) on 60 classes with a frame-level IoU of 0.5. In addition to that, we also evaluate our method on AVA-Kinetics~\cite{li2020ava} dataset, which provides spatio-temporal localized annotations for Kinetics videos.

 \pdfoutput=1
We use PHALP~\cite{rajasegaran2022tracking} to track people in the AVA dataset. PHALP falls into the tracking-by-detection paradigm and uses Mask R-CNN~\cite{he2017mask} for detecting people in the scene. At the training stage, where the bounding box annotations are available only at 1Hz, we use Mask R-CNN detections for the in-between frames and use the ground-truth bounding box for every 30 frames. For validation, we use the bounding boxes used by ~\cite{pan2021actor} and do the same strategy to complete the tracking. We ran, PHALP on Kinetics-400~\cite{kay2017kinetics} and AVA~\cite{gu2018ava}. Both datasets contain over 1 million tracks with an average length of 3.4s and over 100 million detections. In total, we use about 900 hours length of tracks, which is about 40x more than previous works~\cite{kanazawa2019learning}. See Table~\ref{tbl:phalp_tracks} for more details. 

Tracking allows us to train actions densely. Since, we have tokens for each actor at every frame, we can supervise every token by assuming the human action remains the same in a \textit{1 sec} window~\cite{gu2018ava}. First, we pre-train our model on Kinetics-400 dataset~\cite{kay2017kinetics} and AVA~\cite{gu2018ava} dataset. We run MViT~\cite{fan2021multiscale} (pre-trained on MaskFeat~\cite{wei2022masked}) at 1Hz on every track in Kinetics-400 to generate pseudo ground-truth annotations. Every 30 frames will share the same annotations and we train our model end-to-end with binary cross-entropy loss. Then we fine-tune the pretrained model, with \textit{tracks} generated by us, on AVA ground-truth action labels. At inference, we take a track, and randomly sample $N-1$ of other tracks from the same video and pass it through the model. We take an average pooling on the prediction head over a sequence of $12$ frames, and evaluate at the center-frame. For more details on model architecture, hyper-parameters, and training procedure/training-time please see Appendix A1.

\begin{figure*}[!ht]
    \centering
    \includegraphics[width=0.98\textwidth]{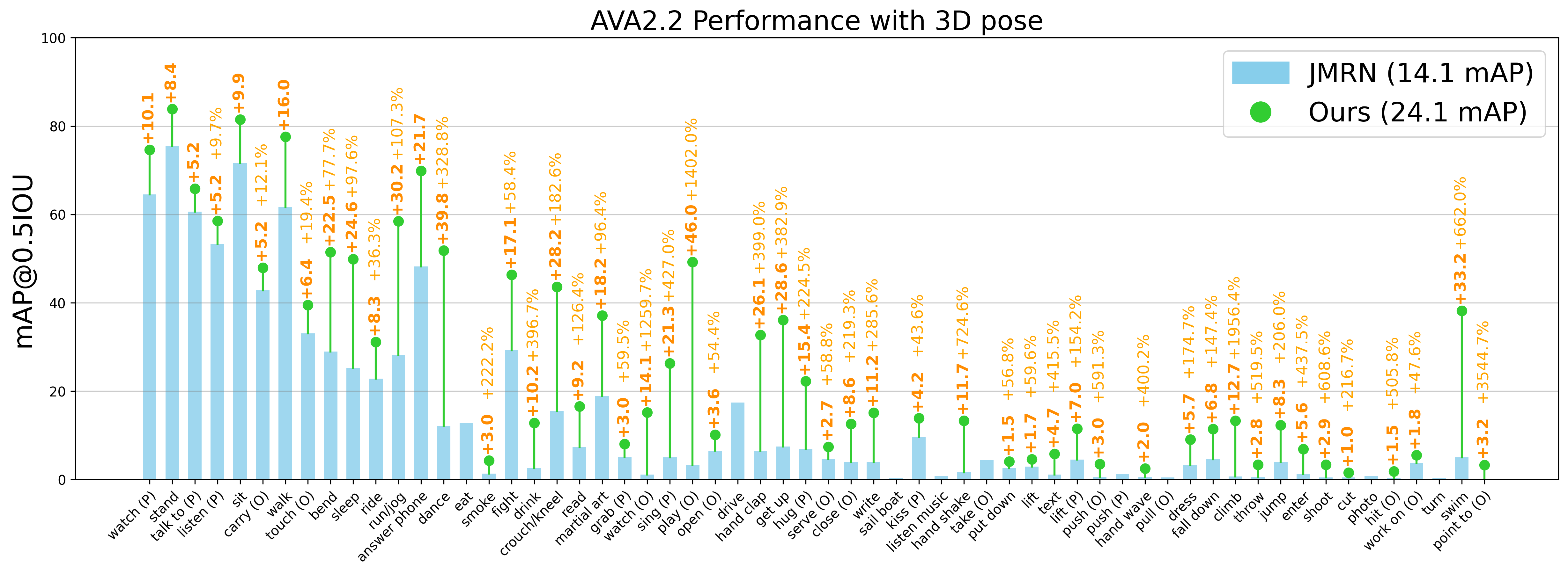}
    \caption{\textbf{Class-wise performance on AVA:} We show the performance of JMRN~\cite{shah2022pose} and \methodnameA \ on 60 AVA classes (average precision and relative gain). For pose based classes such as \textit{standing}, \textit{sitting}, and \textit{walking} our 3D pose model can achieve above 60 mAP average precision performance by only looking at the 3D poses over time. By modeling multiple trajectories as input our model can understand the interactions among people. For example, activities such as \textit{dancing} (\textcolor{teal}{+30.1\%}), \textit{martial art} (\textcolor{teal}{+19.8\%}) and \textit{hugging} (\textcolor{teal}{+62.1\%}) have large relative gains over state-of-the-art pose only model. We only plot the gains if it is above or below 1 mAP.}
    \label{fig:single_ve_multi_pose}
    \vspace{-0.4cm}
\end{figure*}

\begin{table}[!t]
\begin{center}
\small
\vspace{5pt}
\begin{tabular}{l c c c }
\toprule[0.4mm]
Dataset & \# clips & \# tracks & \# bbox \\ \midrule
AVA~\cite{gu2018ava} & 184k & 320k & 32.9m \\
Kinetics~\cite{kay2017kinetics} & 217k & 686k & 71.4m \\ \midrule
Total & 400k & 1m & 104.3m\\
\bottomrule[0.4mm]
\end{tabular}
\end{center}
\vspace{-10pt}
\caption{\textbf{Tracking statistics on AVA~\cite{gu2018ava} and Kinetics-400~\cite{kay2017kinetics}:} We report the number tracks returned by PHALP~\cite{rajasegaran2022tracking} for each datasets (m: million). This results in over 900 hours of tracks, with a mean length of 3.4 seconds (with overlaps). }
\label{tbl:phalp_tracks}
\end{table}

\subsection{Action Recognition with 3D Pose}
\label{sec:experiments_pose_only}

In this section, we discuss the performance of our method on AVA action recognition, when using 3D pose cues, corresponding to Section~\ref{sec:method_pose}. We train our 3D pose model \textbf{\methodnameA}, on Kinetics-400 and AVA datasets. For Kinetics-400 tracks, we use MaskFeat~\cite{wei2022masked} pseudo-ground truth labels and for AVA tracks, we train with ground-truth labels. We train a single person model and a multi-person model to study the interactions of a person over time, and person-person interactions. Our method achieves 24.1 mAP on multi-person (N=5) setting (See Table~\ref{tbl:pose_only}). While this is well below the state-of-the-art performance, this is a first time a 3D model achieves more than 15.6 mAP on AVA dataset. Note that the first reported performance on AVA was 15.6 mAP~\cite{gu2018ava}, and our 3D pose model is already above this baseline.

\begin{table}[!t]
\begin{center}
\small
\begin{tabular}{l c c c c c}
\toprule[0.4mm]
Model & Pose &  OM  & PI & PM & mAP \\ \midrule
PoTion~\cite{choutas2018potion}        & 2D        & - & - & - & 13.1 \\ 
JMRN~\cite{shah2022pose}               & 2D        & 7.1   & 17.2 & 27.6 & 14.1 \\ \midrule 
\methodnameA                                   & 3D (n=1)    &  12.0 & 22.0 & 46.6 & 22.9 \\
\methodnameA                                   & 3D (n=5)    &  13.3 & 25.9 & 48.7 & 24.1 \\
\bottomrule[0.4mm]
\end{tabular}
\end{center}
\vspace{-10pt}
\caption{\textbf{AVA Action Recognition with 3D pose:} We evaluate human-centric representation on AVA dataset~\cite{gu2018ava}. Here \textit{OM} : Object Manipulation, \textit{PI} : Person Interactions, and \textit{PM} : Person Movement. \methodnameA can achieve about 80\% performance of MViT models on person movement tasks without looking at scene information.}
\vspace{-0.2cm}
\label{tbl:pose_only}
\end{table}

We evaluate the performance of our method on three AVA sub-categories (Object Manipulation (\textit{OM}), Person Interactions (\textit{PI}), and Person Movement(\textit{PM})). For the person-movement task, which includes actions such as \textit{running}, \textit{standing}, and \textit{sitting} etc., the 3D pose model achieves 48.7 mAP. In contrast, MaskFeat performance in this sub-category is 58.6 mAP. This shows that the 3D pose model can perform about 80\% good as a strong state-of-the-art model. On the person-person interaction category, our multi-person model achieves a gain of +2.1 mAP compared to the single-person model, showing that the multi-person model was able to capture the person-person interactions. As shown in the Fig~\ref{fig:single_ve_multi_pose}, for person-person interactions classes such as \textit{dancing}, \textit{fighting}, \textit{lifting a person} and \textit{handshaking} etc., the multi-person model performs much better than the current state-of-the-art pose-only models. For example, in \textit{dancing} multi-person model gains +39.8 mAP, and in \textit{hugging} the relative gain is over +200\%. In addition to that, the multi person model has the largest gain compared to the single person model in the person interactions category.

On the other hand, object manipulation has the lowest score among these three tasks. Since we do not model objects explicitly, the model has no information about which object is being manipulated and how it is being associated with the person. However, since some tasks have a unique pose when interacting with objects such as \textit{answering a phone} or \textit{carrying an object}, knowing the pose would help in identifying the action, which results in 13.3 mAP.

\subsection{Actions from Appearance and 3D Pose}

While the 3D pose model can capture about 50\% performance compared to the state-of-the-art methods, it does not reason about the scene context. To model this, we concatenate the human-centric 3D representation with feature vectors from MaskFeat~\cite{wei2022masked} as discussed in Section~\ref{sec:method_pose_apperance}. MaskFeat has a MViT2~\cite{li2021improved} as the backbone and it learns a strong representation about the scene and contextualized appearance. First, we pretrain this model on Kinetics-400~\cite{kay2017kinetics} and AVA~\cite{gu2018ava} datasets, using the pseudo ground truth labels. Then, we fine-tune this model on AVA tracks using the ground-truth action annotation. 

\begin{figure*}[!ht]
    \centering
    \includegraphics[width=0.98\textwidth]{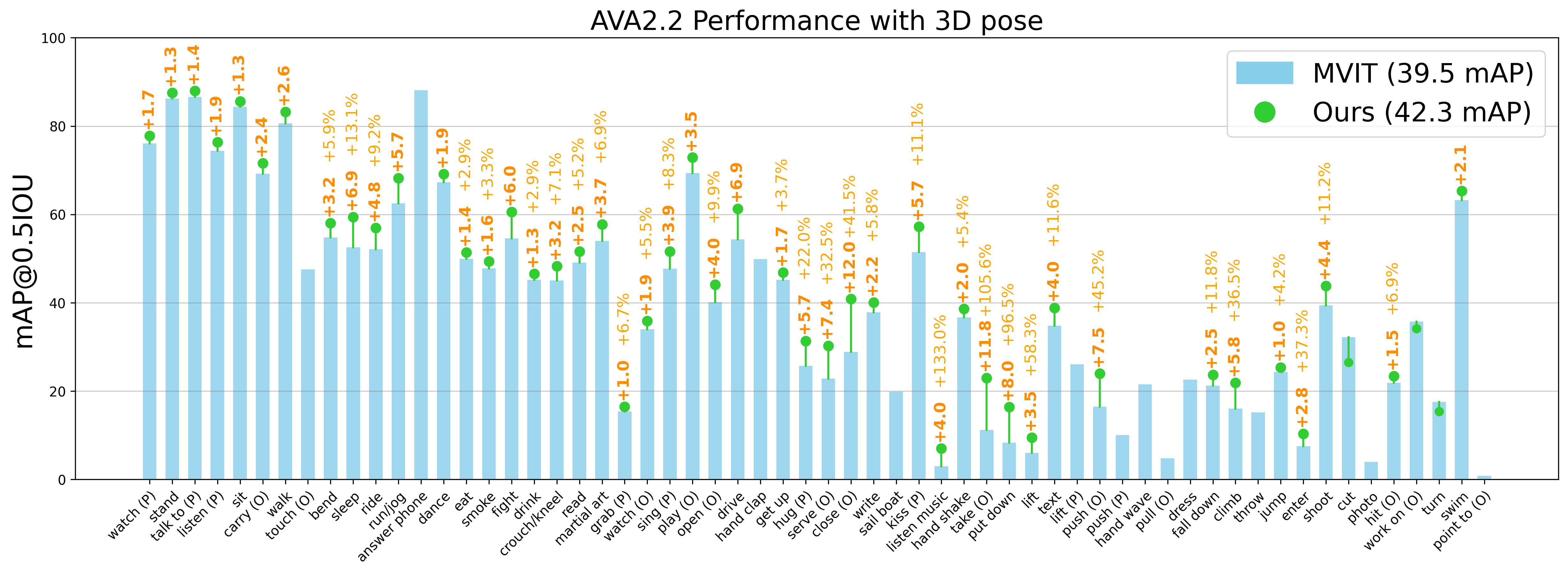}
    \caption{\textbf{Comparison with State-of-the-art methods:} We show class-level performance (average precision and relative gain) of MViT~\cite{fan2021multiscale} (pretrained on MaskFeat~\cite{wei2022masked}) and ours. Our methods achieve better performance compared to MViT on over 50 classes out of 60 classes. Especially, for actions like \textit{running}, \textit{fighting}, \textit{hugging}, and \textit{sleeping} etc., our method achieves over \textbf{+5 mAP}. This shows the benefit of having access to explicit tracks and 3D poses for action recognition. We only plot the gains if it is above or below 1 mAP.}
    \label{fig:results_sota}
\end{figure*}

In Table~\ref{tbl:results_sota} we compare our method with other state-of-the-art methods. Overall our method has a gain of \textbf{+2.8 mAP} compared to Video MAE~\cite{feichtenhofer2022masked, tong2022videomae}. In addition to that if we train with extra annotations from AVA-Kinetics our method achieves \textbf{42.3 mAP}. Figure~\ref{fig:results_sota} show the class-wise performance of our method compared to MaskFeat~\cite{wei2022masked}. Our method overall improves the performance of 56 classes in 60 classes. For some classes (e.g. \textit{fighting}, \textit{hugging}, \textit{climbing}) our method improves the performance by more than +5 mAP. In Table~\ref{tbl:AVA-K} we evaluate our method on AVA-Kinetics~\cite{li2020ava} dataset. Compared to the previous state-of-the-art methods our method has a gain of +1.5 mAP.

\begin{figure*}[!ht]
    \centering
    \includegraphics[width=1\textwidth]{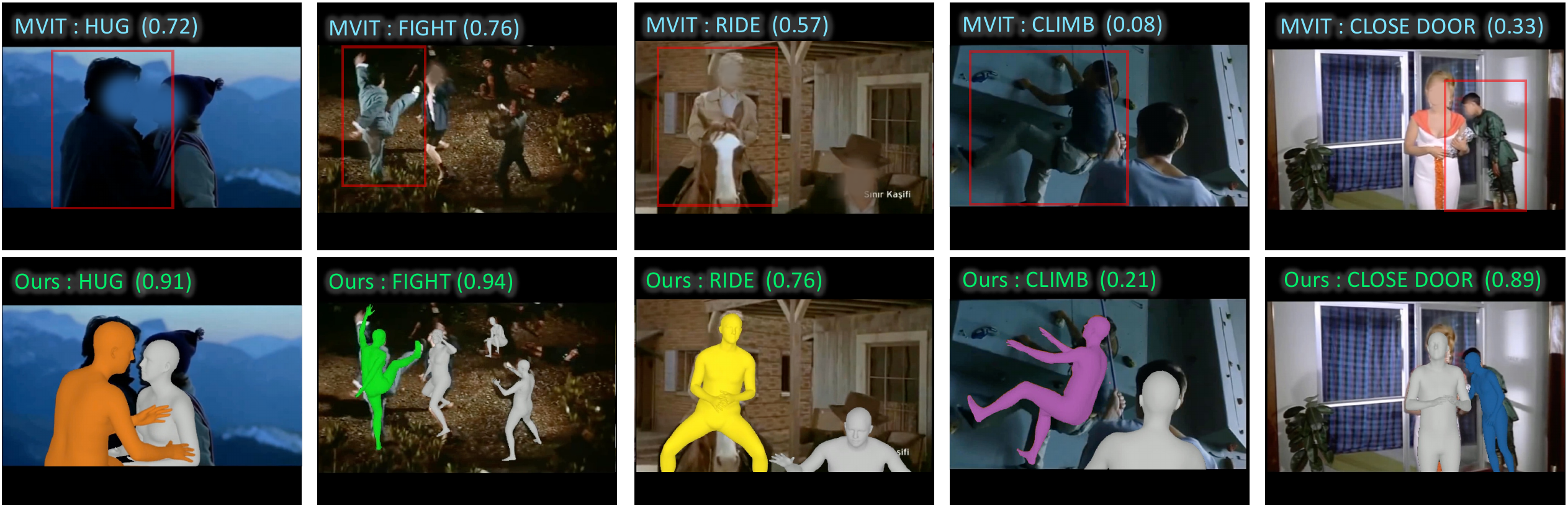}
    \caption{\textbf{Qualitative Results:} 
    We show the predictions from MViT~\cite{fan2021multiscale} and our model on validation samples from AVA v2.2. The person with the colored mesh indicates the person-of-interest for which we recognise the action and the one with the \tikzmarknode[draw,inner sep=2pt,rounded corners,fill=gray!30]{A}{gray mesh} indicates the supporting actors.
    The first two columns demonstrate the benefits of having access to the action-tubes of other people for action prediction. In the first column, the \tikzmarknode[draw,inner sep=2pt,rounded corners,fill=orange!30]{A}{orange person} is very close to the other person with hugging posture, which makes it easy to predict \emph{hugging} with higher probability. Similarly, in the second column, the explicit interaction between the multiple people, and knowing others also fighting increases the confidence for the \emph{fighting} action for the \tikzmarknode[draw,inner sep=2pt,rounded corners,fill=green!30]{A}{green person} over the 2D recognition model. The third and the fourth columns show the benefit of explicitly modeling the 3D pose over time (using tracks) for action recognition. Where the \tikzmarknode[draw,inner sep=2pt,rounded corners,fill=yellow!30]{A}{yellow person} is in riding pose and \tikzmarknode[draw,inner sep=2pt,rounded corners,fill=Thistle!30]{A}{purple person} is looking upwards and legs on a vertical plane. The last column indicates the benefit of representing people with an amodal representation. Here the hand of the \tikzmarknode[draw,inner sep=2pt,rounded corners,fill=RoyalBlue!30]{A}{blue person} is occluded, so the 2D recognition model does not see the action as a whole. However, SMPL meshes are amodal, therefore the hand is still present, which boosts the probability of predicting the action label for \emph{closing the door}.}
    \label{fig:results_qualitative}
\end{figure*}

In Figure~\ref{fig:results_qualitative}, we show qualitative results from MViT~\cite{fan2021multiscale} and our method. As shown in the figure, having explicit access to the tracks of everyone in the scene allow us to make more confident predictions for actions like \emph{hugging} and \emph{fighting}, where it is easy to interpret close interactions. In addition to that, some actions like \emph{riding a horse} and \emph{climbing} can benefit from having access to explicit 3D poses over time. Finally, the amodal nature of 3D meshes also allows us to make better predictions during occlusions.

\begin{table}[!h]
\begin{center}
\small
\begin{tabular}{l c l}
\toprule[0.4mm]
Model & Pretrain & mAP \\ \midrule
SlowFast R101, 8×8~\cite{feichtenhofer2019slowfast}        & \multirow{2}{*}{K400}  & 23.8 \\
MViTv1-B, 64×3~\cite{fan2021multiscale}                            &                                          & 27.3 \\ \midrule
SlowFast 16×8 +NL~\cite{feichtenhofer2019slowfast}       &  \multirow{5}{*}{K600} & 27.5 \\
X3D-XL~\cite{feichtenhofer2020x3d}                                    &                                          & 27.4 \\
MViTv1-B-24, 32×3~\cite{fan2021multiscale}                      &                                           & 28.7 \\
Object Transformer~\cite{wu2021towards}                         &                                           & 31.0 \\
ACAR R101, 8×8 +NL~\cite{pan2021actor}                         &                                          & 31.4 \\ \midrule
ACAR R101, 8×8 +NL~\cite{pan2021actor}                         &  \multirow{1}{*}{K700} & 33.3 \\ \midrule
MViT-L↑312, 40×3~\cite{li2021improved},                           &  IN-21K+K400       & 31.6 \\ 
MaskFeat~\cite{wei2022masked}                                          &  K400                               & 37.5 \\
MaskFeat~\cite{wei2022masked}                                          &  K600                               & 38.8 \\
Video MAE~\cite{feichtenhofer2022masked,tong2022videomae}                                  &  K600                               & 39.3 \\ 
Video MAE~\cite{feichtenhofer2022masked,tong2022videomae}                                  &  K400                               & 39.5 \\  
Hiera~\cite{ryali2023hiera}                                  &  K700                               & 42.3 \\  \midrule
\methodnameB - MViT                                               &  K400                               & 42.6$\ \textcolor{teal}{(+2.8)}$ \\         
\methodnameB - Hiera                                               &  K700                               & 45.1$\ \textcolor{teal}{(+2.5)}$ \\         
\bottomrule[0.4mm]
\end{tabular}
\end{center}
\vspace{-0.2cm}
\caption{\textbf{Comparison with state-of-the-art methods on AVA 2.2:}. Our model uses features from MaskFeat~\cite{wei2022masked} with full crop inference. Compared to Video MAE~\cite{feichtenhofer2022masked, tong2022videomae} our method achieves a gain of \textbf{+2.8 mAP}, and with Heira~\cite{ryali2023hiera} backbone our model achieves 45.1 mAP with a gain of \textbf{+2.5 mAP}.}
\vspace{-5pt}
\label{tbl:results_sota}
\end{table}

\begin{table}[!h]
\begin{center}
\small
\begin{tabular}{l c}
\toprule[0.4mm]
Model & mAP \\ \midrule
SlowFast~\cite{feichtenhofer2019slowfast} & 32.98 \\ \midrule
ACAR~\cite{pan2021actor} & 36.36  \\ \midrule
RM~\cite{feng2021relation} & 37.34 \\ \midrule
\methodnameB+MViT & \textbf{38.91}  \\
\bottomrule[0.4mm]
\end{tabular}
\end{center}
\vspace{-5pt}
\caption{\textbf{Performance on AVA-Kinetics Dataset.} We evaluate the performance of our model on AVA-Kinetics~\cite{li2020ava} using a single model (no ensembles) and compare the performance with previous state-of-the-art single models. }
\vspace{-0.5cm}
\label{tbl:AVA-K}
\end{table}

\subsection{Ablation Experiments}

\paragraph{Effect of tracking:} All the current works on action recognition do not associate people over time, explicitly. They only use the mid-frame bounding box to predict the action. For example, when a person is \textit{running} across the scene from left to right, a feature volume cropped at the mid-frame bounding box is unlikely to contain all the information about the person. However, if we can track this person we could simply know their exact position over time and that would give more localized information to the model to predict the action.

To this end, first, we evaluate MaskFeat~\cite{wei2022masked} with the same detection bounding boxes~\cite{pan2021actor} used in our evaluations, and it results in 40.2 mAP. With this being the baseline for our system, we train a model which only uses MaskFeat features as input, but over time. This way we can measure the effect of tracking in action recognition. Unsurprisingly, as shown in Table~\ref{tbl:main_ablation} when training MaskFeat with tracking, the model performs +1.2 mAP better than the baseline. This clearly shows that the use of tracking is helpful in action recognition. Specifically, having access to the tracks help to localize a person over time, which in return provides a second order signal of how joint angles changes over time. In addition, knowing the identity of each person also gives a discriminative signal between people, which is helpful for learning interactions between people.

\paragraph{Effect of Pose:} The second contribution from our work is to use 3D pose information for action recognition. As discussed in Section~\ref{sec:experiments_pose_only} by only using 3D pose, we can achieve 24.1 mAP on AVA dataset. While it is hard to measure the exact contribution of 3D pose and 2D features, we compare our method with a model trained with only MaskFeat and tracking, where the only difference is the use of 3D pose. As shown in Table~\ref{tbl:main_ablation}, the addition of 3D pose gives a gain of +0.8 mAP. While this is a relatively small gain compared to the use of tracking, we believe with more robust and accurate 3D pose systems, this can be improved.

\begin{table}[!h]
\begin{center}
\small
\vspace{5pt}
\begin{tabular}{c c c c l}
\toprule[0.4mm]
Model & OM & PI & PM & mAP \\ \midrule
MViT                   & 32.2 & 41.1 & 58.6 & 40.2 \\
MViT + Tracking        & 33.4 & 43.0 & 59.3 & 41.4$\ \textcolor{teal}{(+1.2)}$ \\
MViT + Tracking + Pose & 34.4 & 43.9 & 59.9 & 42.3$\ \textcolor{teal}{(+0.9)}$ \\
\bottomrule[0.4mm]
\end{tabular}
\end{center}
\vspace{-10pt}
\caption{\textbf{Ablation on the main components:} We ablate the contribution of tracking and 3D poses using the same detections. First, we only use MViT features over the tracks to evaluate the contribution from tracking. Then we add 3D pose features to study the contribution from 3D pose for action recognition.} 
\vspace{-0.4cm}
\label{tbl:main_ablation}
\end{table}

\subsection{Implementation details}
In both the pose model and pose+appearance model, we use the same vanilla transformer architecture~\cite{vaswani2017attention} with 16 layers and 16 heads. For both models the embedding dimension is $512$. We train with 0.4 mask ratio and at test time use the same mask token to in-fill the missing detections. The output token from the transformer is passed to a linear layer to predict the AVA action labels. We pre-train our model on kinetics for 30 epochs with MViT~\cite{fan2021multiscale} predictions as pseudo-supervision and then fine-tune on AVA with AVA ground truth labels for few epochs. We train our models with AdamW~\cite{loshchilov2017decoupled} with base learning rate of $0.001$ and $\textrm{betas}=(0.9,0.95)$. We use cosine annealing scheduling with a linear warm-up. For additional details please see the Appendix.

\subsection{Hiera backbone results}
\label{sec:hiera_results}

In this section, we show the class-wise performance when Hiera~\cite{ryali2023hiera} is used to extract contextualized appearance features. Our model achieves \textbf{2.5 mAP} gain over the Hiera baseline. Overall over method achieves 45.1 mAP and significant gains on multiple action classes. Table~\ref{tbl:hiera_table} shows task wise performance of the Hiera model and the performance of LART. In all three categories, LART performs much better than the baseline model, and LART shows a significant gain in person-interaction tasks.

\begin{table}[!h]
\begin{center}
\small
\begin{tabular}{l c c c c}
\toprule[0.4mm]
Model &  OM  & PI & PM & mAP \\ \midrule
MViT~\cite{wei2022masked}              & 32.2 & 41.1 & 58.6 & 40.5 \\ 
\methodnameB+MViT                      & 34.5 & 44.0 & 59.9 & 42.3 \\ \midrule 
Hiera~\cite{ryali2023hiera}            & 34.5 & 43.0 & 62.3 & 42.6 \\
\methodnameB+Hiera                     & 37.3 & 45.7 & 63.8 & 45.1 \\
\bottomrule[0.4mm]
\end{tabular}
\end{center}
\vspace{-10pt}
\caption{\textbf{Effect of Backbones:} AVA dataset~\cite{gu2018ava} contains \textit{OM} : Object Manipulation, \textit{PI} : Person Interactions, and \textit{PM} : Person Movement tasks. We compare the action detection performance of the backbones (MViT~\cite{wei2022masked} and Hiera~\cite{ryali2023hiera}) and the performance when combined with LART. }
\vspace{-0.2cm}
\label{tbl:hiera_table}
\end{table}

\begin{figure*}[!ht]
    \centering
    \includegraphics[width=0.98\textwidth]{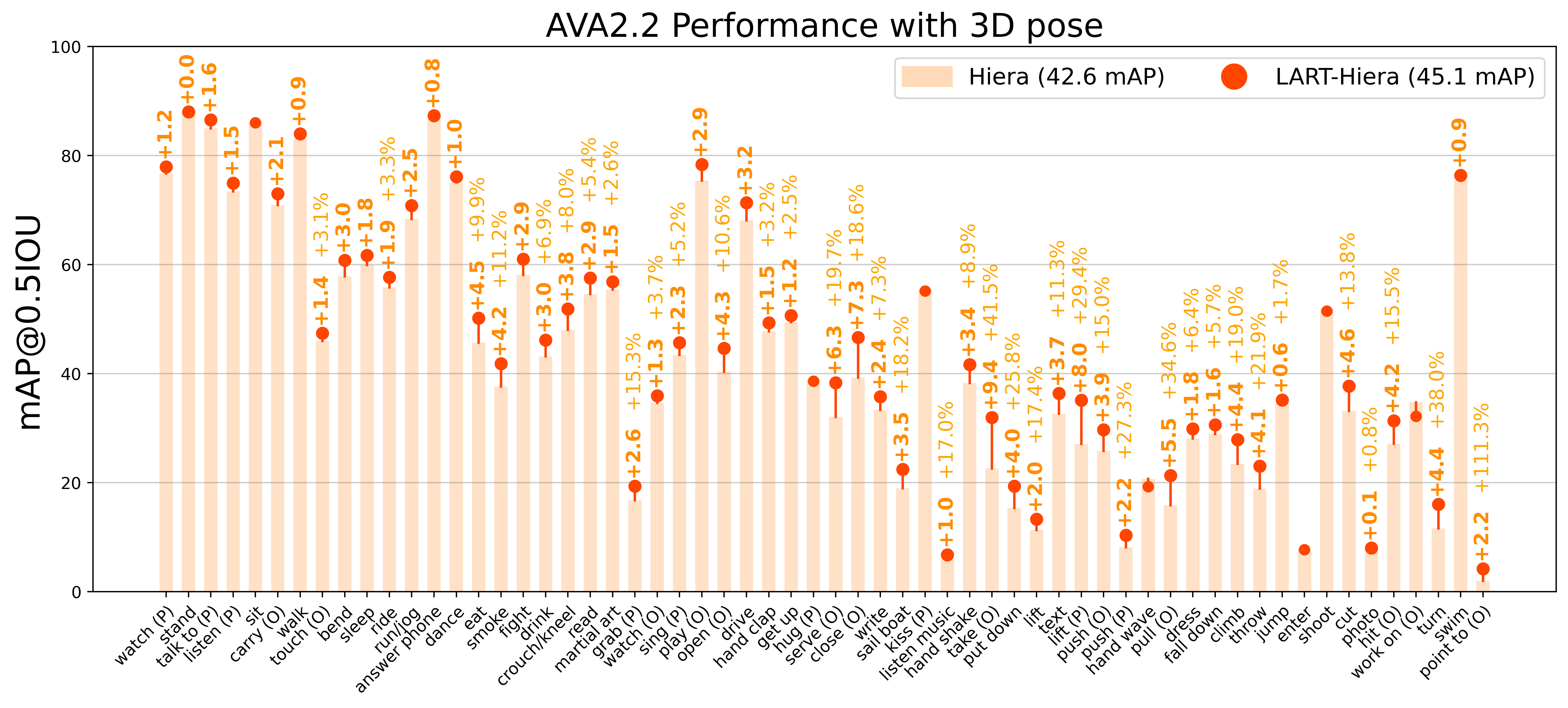}
    \caption{\textbf{Comparison with Hiera~\cite{ryali2023hiera}:} We show class-level performance (average precision and relative gain) of Hiera~\cite{ryali2023hiera} (pre-trained on using MAE~\cite{he2021masked}) and ours. Our methods achieve better performance compared to Hiera in 53 classes out of 60 classes. Especially, for actions involving object interactions, and pose changes such as running. fighting, our method achieves over 10\% relative gain on mean average precision. This shows the benefit of having access to explicit tracks and 3D poses for action recognition. We only plot the gains if it is above or below 1 mAP.}
    \label{fig:results_sota_mvit}
\end{figure*}

 \pdfoutput=1
\section{Conclusion}
\label{sec:discussion}

In this paper, we investigated the benefits of 3D tracking and pose for the task of human action recognition.
By leveraging a state-of-the-art method for person tracking, PHALP~\cite{rajasegaran2022tracking}, we trained a transformer model that takes as input tokens the state of the person at every time instance.
We investigated two design choices for the content of the token.
First, when using information about the 3D pose of the person, we outperform previous baselines that rely on pose information for action recognition by 8.2 mAP on the AVA v2.2 dataset.
Then, we also proposed fusing the pose information with contextualized appearance information coming from a typical action recognition backbone~\cite{fan2021multiscale} applied over the tracklet trajectory.
With this model, we improved upon the previous state-of-the-art on AVA v2.2 by 2.8 mAP. With better backbones such as Heira~\cite{ryali2023hiera} LART achieves 45.1 mAP on AVA v2.2 action detection tasks.
There are many avenues for future work and further improvements for action recognition.
For example, one could achieve better performance for more fine-grained tasks by more expressive 3D reconstruction of the human body (\eg, using the SMPL-X model~\cite{pavlakos2019expressive} to capture also the hands), and by explicit modeling of the objects in the scene (potentially by extending the ``tubes'' idea to objects).

\noindent \textbf{Acknowledgements:} This work was supported by the FAIR-BAIR program as well as ONR MURI (N00014-21-1-2801). We thank Shubham Goel, for helpful discussions.

\balance
{\small
\bibliographystyle{ieee_fullname}
\bibliography{egbib}
}

\newpage
\appendix
 \pdfoutput=1
\section{Supplementary Materials}
\label{sec:supmat_intro}

In this document, we provide additional details about our method that were not included in the main manuscript, due to space constraints. We include additional experiments and implementation details about our approach (Sections 2 \& 3), we provide more details about the experiments of our paper (Sections 4) and we include the training configurations for Kinetics-400 and AVA for reproducibility.

\section{Additional Results}
\label{sec:supmat_results}

\subsection{Modeling Multiple People in the Scene}
\begin{figure}[!h]
    \centering
    \includegraphics[width=0.98\linewidth]{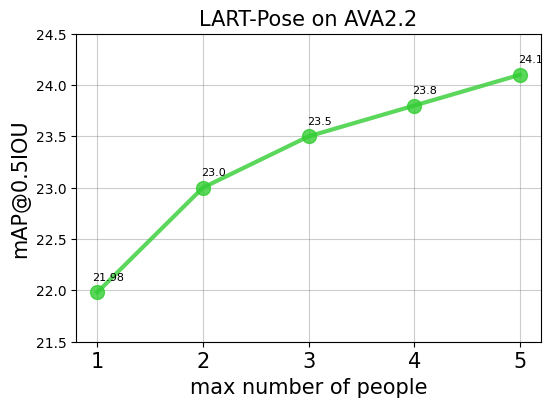}
    \caption{\textbf{\methodnameA \ performance with number of people in the scene:} We show the performance of \methodnameA \ on various $n$ (maximum number of people the model sees in every frame). This plot shows that as we increase the number of people the model can see while reasoning about the person of interest, the performance increase monotonically. While our multi-person model is very simple; we just input additional tokens for each other people in the scene, the model is able to understand these interactions from the large scale training data.  }
    \label{fig:NUM_PEOPLE}
\end{figure}

As discussed in Section~\ref{sec:method} our model take any number of people (tracks) given enough memory. Even though we do simple random sampling to find supporting actors in the scene, knowing this additional context of where other people are located and what they are doing could be a strong signal to predict what the person of interest is doing. We train multiple models, by varying the maximum context for people $(n \in [1,2,3,4,5])$. When $n=1$, the \methodnameB only sees the person-of-interest and the information about the scene and other people are fed through the contextualized appearance vector. However, with larger $n$, other people's poses, locations and appearance are explicitly given to the model. As shown in the Figure~\ref{fig:NUM_PEOPLE} as we increase the \methodnameA \  model's people-context, the performance on AVA dataset increases monotonically, and starts saturating at $n>4$.

This observation highlights the benefits of modeling multiple people in the scene and learning their interactions jointly. However, it is only possible to achieve these benefits if we can track every person in the scene. We observed saturation at $n=5$ with simple random sampling, but this observation may be heavily biased for the AVA dataset. Most movies have few characters, and the biases in the way movies are captured (e.g., close-ups for kissing and hugging scenes) will have an impact on these results. It's important to note that all of these models have the same number of parameters and were trained for the same amount of time (30 epochs on the same dataset).

In summary, our findings suggest that modeling multi-person interactions can significantly improve the performance of action recognition models, particularly for actions that involve close person-person interactions or group activities. However, the saturation point may vary depending on the dataset and biases in the way scenes are captured. 

\subsection{Single person vs Multi person Results}

\begin{figure*}[!ht]
    \centering
    \includegraphics[width=0.98\textwidth]{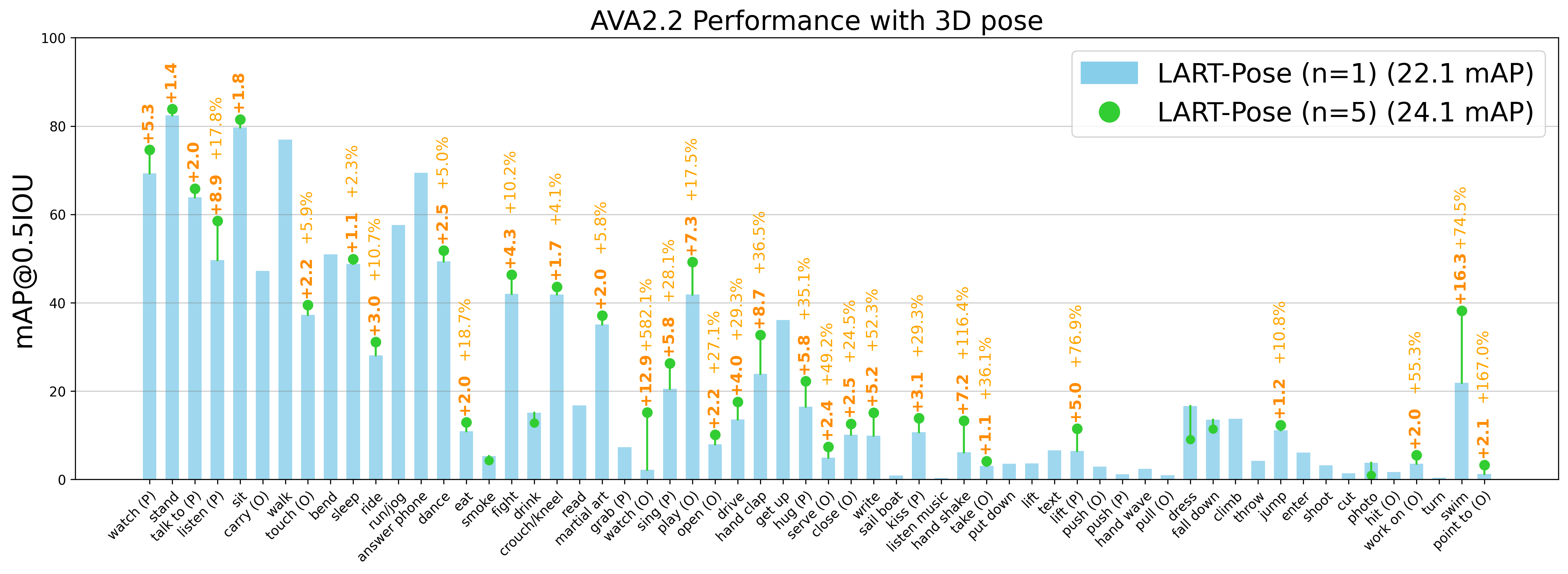}
    \caption{\textbf{Class-wise performance on AVA:} We compare \methodnameA on single person $(n=1)$ and multi-person $(n=5)$ setting. Our multi-person model outperforms single person model on over 50 classes and on some person-person interaction classes multi-person model has a relative gain of about \textbf{10\%}. }
    \label{fig:1vs5}
    \vspace{-0.2cm}
\end{figure*}

In the previous section, we discussed the benefits of modeling multi-person interactions. In this section, we will study how much performance gain can be achieved from a single person model to a multi-person model. We compare the performance of \methodnameA \ (n=1) and \methodnameA \ (n=5) in Figure~\ref{fig:1vs5}. The multi-person model has an overall gain of \textbf{2.0 mAP} over the single person model.

Upon closer examination of the performance of each class, it becomes apparent that classes involving close person-person interactions benefited greatly from multi-person training. For example, actions such as \emph{hugging}, \emph{kissing}, \emph{handshaking}, \emph{lifting a person}, and \emph{listening to a person} have improved by over \textbf{10\%} relative to the single person model. These interactions occur at very close proximity, and having explicit knowledge of the other person's 3D location, pose, and action would aid in identifying the actions of the main actor.

In addition to the close interaction actions, there are group actions that would also benefit from using the context of other people's actions to reason about the action of the person of interest. For example, \emph{dancing} and \emph{swimming} are typically group activities, and knowing what others are doing is a good signal to infer the action.

From this multi-person model, over 50 classes have gained over the single person model, and over 30 classes have gained over \textbf{1 mAP}. However, as mentioned in the previous section regarding biases in datasets, these results may vary slightly for different types of datasets, such as sports datasets. For example, in a sports scene, it may be necessary to look at more than 5 people to recognize the action of the player, and the way sports scenes are shot is significantly different from movies.

Overall, our findings demonstrate that modeling multi-person interactions can significantly improve the performance of action recognition models, particularly for actions that involve close person-person interactions or group activities.

\section{Implementation details}
\label{sec:supmat_implementaion}
Our complete system for action recognition by tracking integrates multiple sub-systems to combine the recent advancements in 2D Detection, Recognition, 3D Reconstruction, as well as Tracking. We can break the overall pipeline into two parts: \textbf{a)} frames-to-entities and \textbf{b)} entities-to-action. 

The first part is to lift entities from frames (here, we consider entities=people). For this, we use a state-of-the-art tracking algorithm, PHALP~\cite{rajasegaran2022tracking}. The first step of PHALP is to detect people in each frame using Mask R-CNN~\cite{he2017mask}. We used Detectron2's~\cite{wu2019detectron2} \href{https://github.com/facebookresearch/detectron2/blob/main/configs/new_baselines/mask_rcnn_regnety_4gf_dds_FPN_400ep_LSJ.py}{new baseline models} trained with Simple Copy-Paste Data Augmentation~\cite{ghiasi2021simple} with a RegNet-4gf~\cite{radosavovic2020designing} backbone for the detection task. After detecting people in each frame, PHALP uses HMR~\cite{kanazawa2018end, rajasegaran2021tracking, 4DHUMANS} to reconstruct each person in 3D. 
Then, the future location, pose, and appearance of each person are predicted for solving association. PHALP uses Hungarian matching to solve the associations between 3D detections and 3D predictions. Finally, a set of tracks will be returned from the tracking which gives us access to entities (people) over time.

For the second part of our paper, we collect tracks and use them to train a transformer model for action recognition. More details on the network architecture and training and inference protocols will be discussed in the following sections. This part of our paper is a crucial component of our approach to using transformer models for action recognition from 3D tracks.

\subsection{PHALP tracklets}
\label{sec:supmat_phalp}
In this work, every person in both Kinetics-400~\cite{kay2017kinetics} and AVA~\cite{gu2018ava} is tracked. For this, we used the recently proposed 3D tracking algorithm PHALP~\cite{rajasegaran2022tracking}. PHALP allows us to track people in the wild very robustly and gives their 3D representations. However, the ground-truth action annotations for AVA are given as bounding boxes at 1 Hz frequency. On a side note, we do not use the ground truth tracking annotations in AVA dataset, which is also only available at 1 Hz. First, we use the PHALP detection model (\eg, Mask R-CNN) to detect humans in the video, whenever a frame does not have ground-truth annotations. If the frame indeed has an annotation, we take the ground-truth bounding boxes as granted and bypass Mask R-CNN detections. Since AVA only has bounding box annotations, and PHALP~\cite{rajasegaran2022tracking} requires bounding boxes and masks, we use Detectron2~\cite{wu2019detectron2} to extract masks from bounding boxes with Mask R-CNN. For the validation set, we used the detections from ACAR~\cite{pan2021actor}, which are also only available every 30 frames. Therefore, we used a similar strategy to get tracks from bounding boxes available at 1 Hz. For Kinetics, we run PHALP tracking for the whole sequence, which is typically 10s clips. However, since AVA is much longer than Kinetics (15 min), we run the tracker for 4-second windows, centered around the evaluation frame.

\subsection{Architecture details}
\label{sec:supmat_architecture}In all of our experiments, we use a vanilla transformer~\cite{vaswani2017attention} architecture with 16 layers and width of 512. Each layer has 16 self-attention heads followed by layer-norm~\cite{ba2016layer}, and a 2-layer MLP followed by layer-norm. We train all the models with a maximum sequence length of 128 frames per person. In other words, every tracklet is trimmed to have a sequence length of 128 frames. The only data augmentation we use is choosing the starting point of the sequence for random trimming. The transformer blocks are followed by a linear layer that predicts AVA action classes. We train all our models with binary cross-entropy loss.

At training time, we use two types of attention masking. First, since the tracklets are not always continuous due to occlusion and missing detections, we mask the corresponding self-attention of these tokens completely. The loss is not applied to these tokens and this part of the tracklet has no effect on training. The second type of masking is done to simulate these kinds of missing detections at test time. We randomly choose a small number of tokens (based on mask ratio), and replace the person-vector with a learnable mask-token. At the self-attention layer, attention is masked such that these masked-tokens will attend other tokens but other tokens will not attend the masked-tokens. Unlike, the first type of masking, we apply loss on these masked token predictions, since if there is a detection available, then there will be a pseudo-ground truth or ground truth label available for training.

At inference time, we do not do any attention-masking. However, there will be some tracklets with discontinuous detections. At these locations, we use the learned masked-token to infill the predictions for the tracklets. Since we are predicting action labels densely for each frame, we take an average pooling of 12 tokens centered around the annotated detection to minimize the gap between human annotations and model predictions.

\subsubsection{Action with 3D Pose}
In this subsection, we discuss the network architecture used for recognizing action only with 3D pose information over time. The 3D pose has 226 parameters: 207 ($23\times3\times3$) parameters for joint angles, 9 for the global orientation of the person, and 10 for the body shape. In addition to this, the 3D translation of the person in the camera frame is represented by 3 parameters. Overall, in this system, a person-vector has a dimension of 229. This vector is encoded by an MLP with two hidden layers to project this to a 256-dimensional vector. The projected person-vector is then passed to the transformer. We also use the three types of positional encodings for time, track, and space as discussed in the main manuscript (Section 3.1).

\subsubsection{Action with 3D Pose and Appearance}
To encode a strong contextualized appearance feature, we used MViT~\cite{fan2021multiscale} pretrained with MaskFeat~\cite{wei2022masked}. The MViT model for AVA takes a sequence of frames and a mid-frame bounding box to predict the action label of the person of interest (this is a classical example of the Eulerian way of predicting action). In this paper, we use an MViT-L 40$\times$3 model that takes a 4-second clip and samples 40 frames with a temporal stride of 3 frames as the input and a bounding box of a person at the mid-frame. This gives a $1152$-dimensional feature vector before the linear layer in the MViT classifier. We use this $1152$-dimensional feature vector as our contextualized appearance feature and encode it into a $256$ dimensional vector by an MLP with two hidden layers. Now, we have a pose vector (256 dim, from the previous section) and an appearance vector (256 dim). We concatenate these two vectors to build our person-vector for 3D pose with appearance, and the final 512-dimensional vector is passed to the transformer.

\subsection{Training recipe}
\label{sec:supmat_training}

As discussed in Section 3 of the main manuscript, we first pretrain our method on Kinetics-400 dataset, using the tracklets obtained from PHALP~\cite{rajasegaran2022tracking}. Each of these tracklets contains a detection at every frame unless the person is occluded or is not detected due to failure of the detection system. We provide these detection bounding boxes as input to the MViT~\cite{fan2021multiscale} model and generate pseudo ground truth action labels for the tracklets. Once the labels are generated, we train our model end-to-end, with tracklets as inputs and the action labels as outputs. We use the training configurations in Table~\ref{tbl:supmat_recipe} for pretraining the model on Kinetics-400 tracklets. Once the model is pretrained on Kinetics tracklets, we fine-tune the model on AVA tracklets (generated by PHALP) with ground truth action labels. Finally, during the fine-tuning stage, we apply layer-wise decay~\cite{clark2020electra} and drop path~\cite{huang2016deep}.

\begin{table}[!h]
\begin{center}
\small
\begin{tabular}{l | c c}
\toprule[0.4mm]
Configs & Kinetics-400 & AVA \\ \midrule
optimizer & \multicolumn{2}{c}{AdamW~\cite{loshchilov2017decoupled}} \\
optimizer momentum  & \multicolumn{2}{c}{$\beta_1, \beta_2=0.9, 0.95$} \\
weight decay & \multicolumn{2}{c}{0.05}\\ 
learning rate schedule & \multicolumn{2}{c}{cosine decay~\cite{loshchilov2016sgdr}}\\ 
warmup epochs & \multicolumn{2}{c}{5}\\ 
drop out & \multicolumn{2}{c}{0.1}\\ 
base learning rate & 1e-3 & 1e-3 \\ 
layer-wise decay~\cite{clark2020electra} & - & 0.9 \\
batch size & 64 & 64 \\ 
training epochs & 30 & 30 \\ 
drop path~\cite{huang2016deep} & - & 0.1 \\
mask ratio & 0.4 & 0.0 \\
\bottomrule[0.4mm]
\end{tabular}
\end{center}
\caption{\textbf{Training Configurations:} We report the training configurations used from training our models on Kinetics-400 and AVA datasets. }
\label{tbl:supmat_recipe}
\end{table}

\end{document}